\documentclass{sig-alternate-05-2015}
\usepackage{subfigure}

\begin{document}






\conferenceinfo{DEEP LEARNING SECURITY WORKSHOP}{'17 SINGAPORE}

\title{On Lyapunov exponents and adversarial perturbations}

\numberofauthors{3} 
%
\author{
\alignauthor
Vinay Uday Prabhu\titlenote{Primary author}\\
       \affaddr{UnifyID Inc}\\
       \affaddr{San Francisco, 94107}\\
       \email{vinay@unify.id}
\alignauthor
Nishant Desai\\
	\affaddr{UnifyID Inc}\\
	\affaddr{San Francisco, 94107}\\
       \email{nishant@unify.id}
\alignauthor
John Whaley\\
       \affaddr{UnifyID Inc}\\
       \affaddr{San Francisco, 94107}\\
       \email{john@unify.id}
}

\maketitle
\begin{abstract}
In this paper, we would like to disseminate a serendipitous discovery involving Lyapunov exponents of a 1-D time series and their use in serving as a filtering defense tool against a specific kind of deep adversarial perturbation. To this end, we use the state-of-the-art CleverHans library to generate adversarial perturbations against a standard  Convolutional Neural Network (CNN) architecture trained on the MNIST as well as the Fashion-MNIST datasets. We empirically demonstrate how the Lyapunov exponents computed on the flattened 1-D vector representations of the images served as highly discriminative features that could be to pre-classify images as adversarial or legitimate before feeding the image into the CNN for classification. We also explore the issue of possible false-alarms when the input images are noisy in a non-adversarial sense.
\end{abstract}

\printccsdesc






\section{Background on defenses against adversarial attacks}
In the recent past, a plethora of defenses against adversarial attacks have been proposed. These include \textit{SafetyNet} \cite{defense_safetynet}, adversarial training \cite{Szegedy_intriguing}, label smoothing \cite{defense_lab_smooth},  defensive distillation \cite{defense_distillation} and feature-squeezing \cite{defense_fsqueezing_1,defense_fsqueezing_2} to name a few. There is also an ongoing Kaggle contest\cite{kaggle_defense} underway for exploring novel defenses against adversarial attacks.
\\ As evinced by the recent spurt in the papers written on this topic, most defenses proposed are \textit{quelled} by a novel attack that exploits some weakness in the defense. In \cite{defense_ensemble}, the authors queried if one could concoct a strong defense by combining 
multiple defenses and showed that an ensemble of weak defenses was not sufficient in providing strong defense against adversarial examples that they were able to craft. In the accompanying  blog\footnote{\url{http://www.cleverhans.io/security/privacy/ml/2017/02/15/why-attacking-machine-learning-is-easier-than-defending-it.html}} associated with \cite{Cleverhans}, the authors Goodfellow and Papernot posit that \textit{Adversarial examples are hard to defend against because it is hard to construct a theoretical model of the adversarial example crafting process} and contend with the idea if \textit{attacking machine learning easier than defending it?}
\\With this background, we shall now look more closely at a specific type of defense and motivate the relevance of our method within this framework.
\subsection{The pre-detector based defenses}
One prominent approach that emerges in the literature of adversarial defenses is that of crafting pre-detection and filtering systems that flag inputs that might be potentially adversarial. In \cite{defense_stats}, the authors posit that adversarial examples are not drawn from the same distribution as the legitimate samples and can thus be detected using statistical tests. In \cite{defense_filter_1,defense_filter_2}, the authors train a separate binary classifier to first classify any input image as legitimate or adversarial and then perform inference on the \textit{passed} images.  In approaches such as \cite{defense_artifacts}, the authors assume that  DNNs classify accurately only near the small manifold of training data and that the synthetic adversarial samples
do not lie on the data manifold. They applying dropout at test time to ascertain the confidence of adversariality of the input image.
\\In this paper, we would like to disseminate a model-agnostic approach towards adversarial defense that is dependent purely on the \textit{quasi-time-series statistics} of the input images that was discovered in a rather serendipitous fashion. The goal is to not present the method we propose as a fool-proof adversarial filter, but to instead draw the attention of the DNN and CV communities towards this chance discovery that we feel is worthy of further inquiry. 
\\In order to facilitate reproducibility of results and effective criticism, we have duly open-sourced the implementation as a well annotated jupyter-notebook shared at the following location: \url{https://github.com/vinayprabhu/Lyapunov_defense}
\section{Procedure followed}
\subsection{A quick introduction to Lyapunov exponents}
For a given a scalar time series $\{ x_t; t=1,...,n\}$ whose time evolution is assumed to be modeled by a differentiable dynamical system (in a phase-space of possibly infinite dimensions), we can define Lyapunov exponents corresponding to the \textit{large-time} behavior of the system. These \textit{characteristic exponents} numerically quantify the sensitivity to the initial conditions and emanate from the Ergodic theory of differentiable dynamical systems, introduced by Eckmann and Ruelle in \cite{eckmann1985ergodic,lyap_e}. Simply put, if the initial state of a time series is slightly perturbed,  the characteristic exponent (or Lyapunov exponents) represent the exponential rate at which the perturbation increases (or decreases) with time.
\\For a detailed understanding of these characteristic exponents, we would like to refer the user to \cite{eckmann1985ergodic,lyap_e}.
\\ Given a finite-length finite-precision time series sequence, the 3 stage \texttt{Embed-Tangent maps-QR decomposition} based recipe introduced in \cite{lyap_e} can be used to numerically compute the Lyapunov exponents. This numerical recipe is implemented in many time-series analysis packages such as \cite{nolds}.
\\The background of the serendipitous discovery was that we were investigating usage of metrics from time-series analysis literature to identify adversarial attacks on 1-D axis-wise mobile-phone motion sensor data ;example, Accelerometer and gyroscope (see \cite{vinay_adv_imu}) and we realized that this procedure could be easily extended to vectorized images (or  \textit{flattened} images viewed as discrete time series indexed by the pixel location. That is, given a normalized image, $\mathbf{X} \in [0,1]^{n \times n}$, we would have the flattened time-series representation $x \in [0,1]^{n^2}$ as simply, 
\begin{equation}
x_{in+j}= \mathbf{X}_{i,j}; \ i,j=0,...n-1;
\end{equation}
\section{Experiments}

In this paper, we used the MNIST \cite{mnist} dataset. In the appendix, we showcase similar results obtained with the Fashion-MNIST dataset \cite{fmnist} as well. The rest of this section details the different stages involved in our experimentation.

\subsection{Generating targeted adversarial perturbations}
\begin{figure*}[!t]
\centering
\subfigure[Image grid with legitimate MNIST images and their adversarial counterparts]{
\includegraphics[width=3.45 in, height=3in] {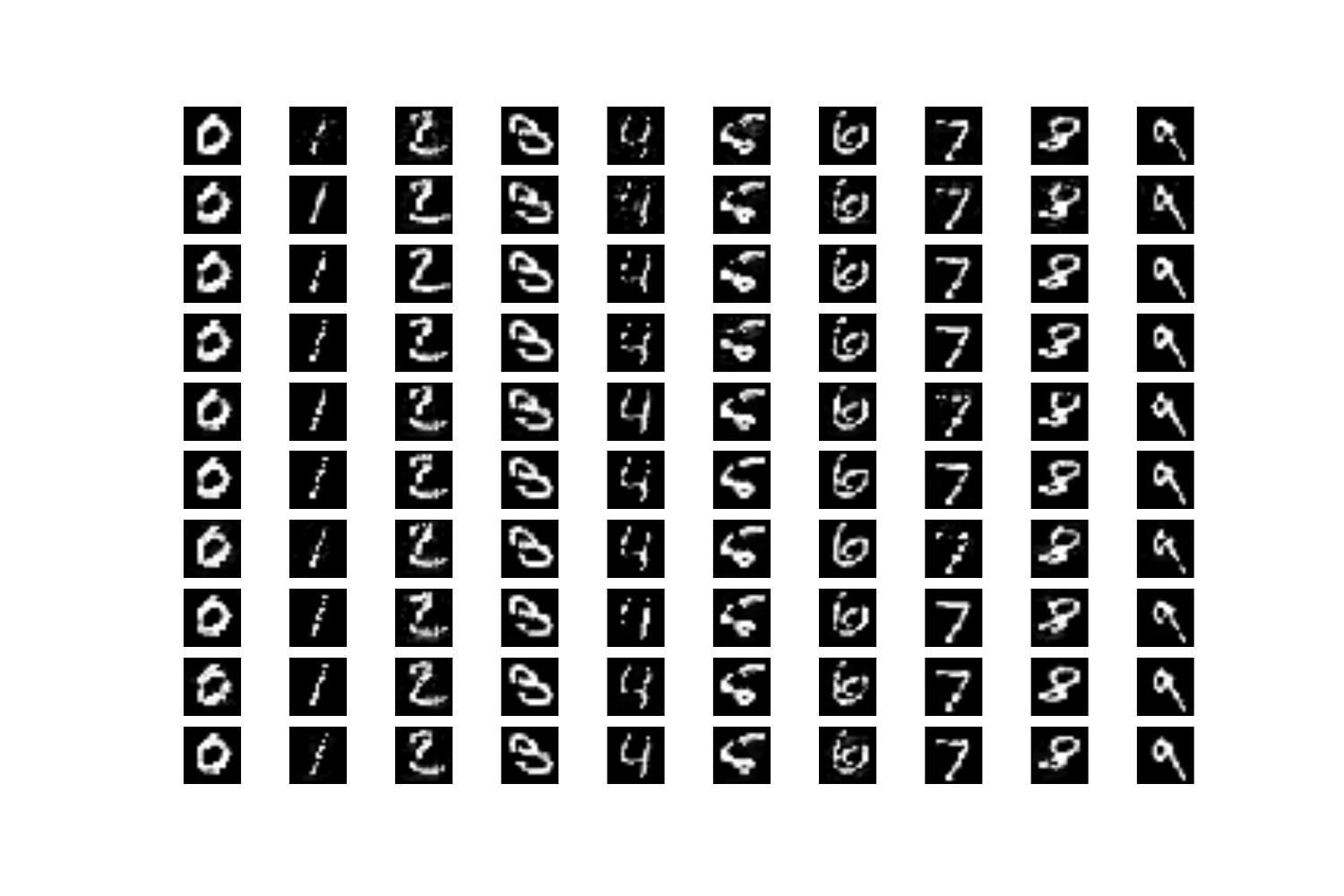}
    \label{cw_mnist}
}
\subfigure[Norm distances for the images in the grid (between legitimate F-MNIST images and their adversarial counterparts)]{
\includegraphics[width=3.25 in, height=2.8 in] {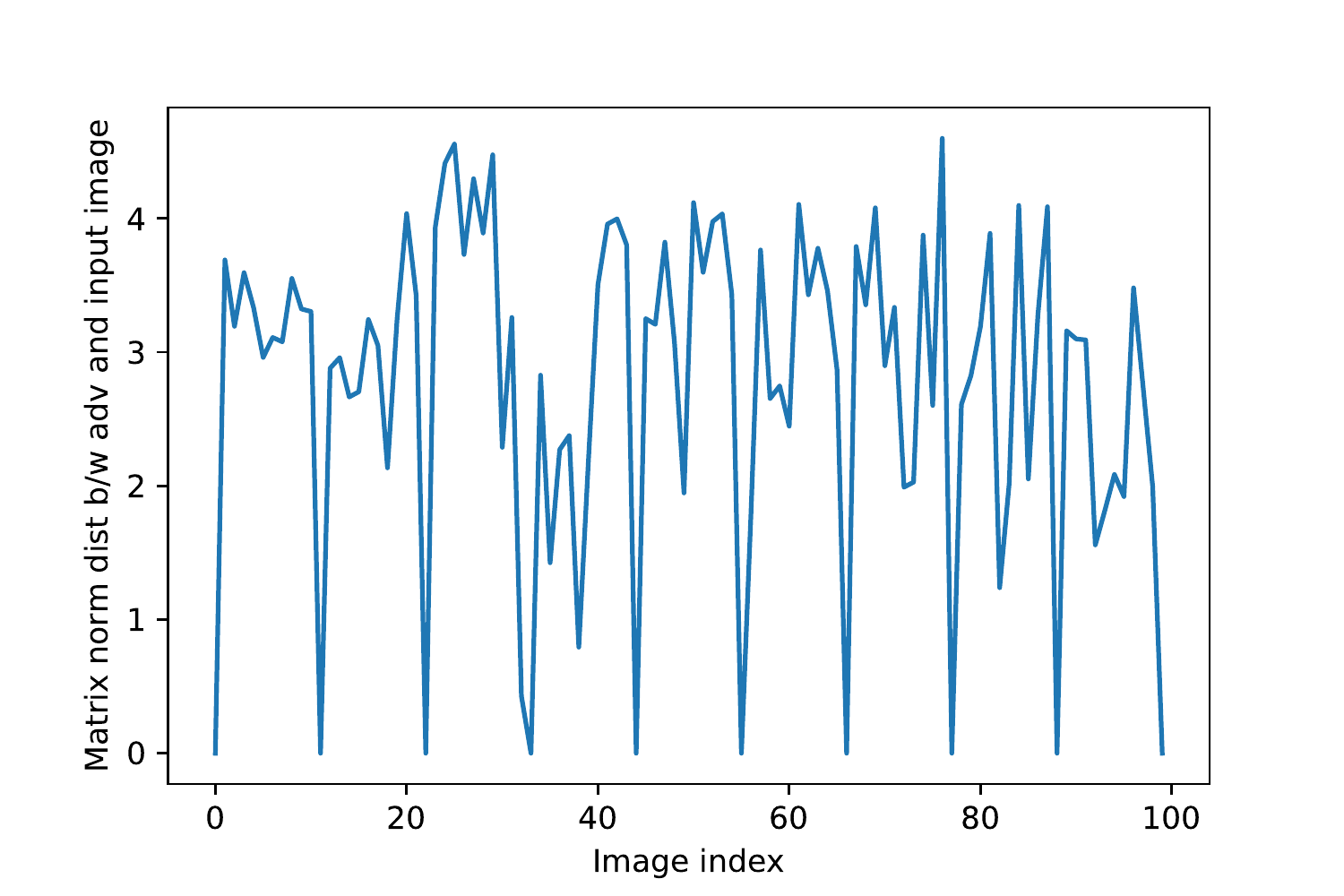}
    \label{dist_cw_mnist}
}
\caption[Optional caption for list of figures]{In the left sub-figure above, the $[i,j]^{th}$ grid image is an image belong to the $j^{th}$ class adversarially perturbed to be classified as belonging to the $i^{th}$ class}
\label{grid}
\end{figure*}

We targeted 10 random samples from the MNIST dataset (1 belong to each class/digit) using the Carlini-Wagner-$l_2$ attack (\cite{cw_l2}) in the targeted mode implemented in the CleverHans library (\cite{Cleverhans}) with the following parameterization.
\begin{verbatim}
    cw_params = {'binary_search_steps': 1,
             'y_target': adv_ys,
             'max_iterations': attack_iterations,
             'learning_rate': 0.1,
             'batch_size': source_samples * nb_classes,
             'initial_const': 10}

\end{verbatim}
The resultant image grid containing the images along with the adversarially perturbed counterparts are as shown in fig \ref{cw_mnist}. Fig \ref{dist_cw_mnist} showcases the norm-distance between the images and their adversarial examples.

\subsection{Computing Lyapunov exponents of the flattened images}
\begin{figure*}[ht]
\centering
\includegraphics[width=7 in, height=8 in]{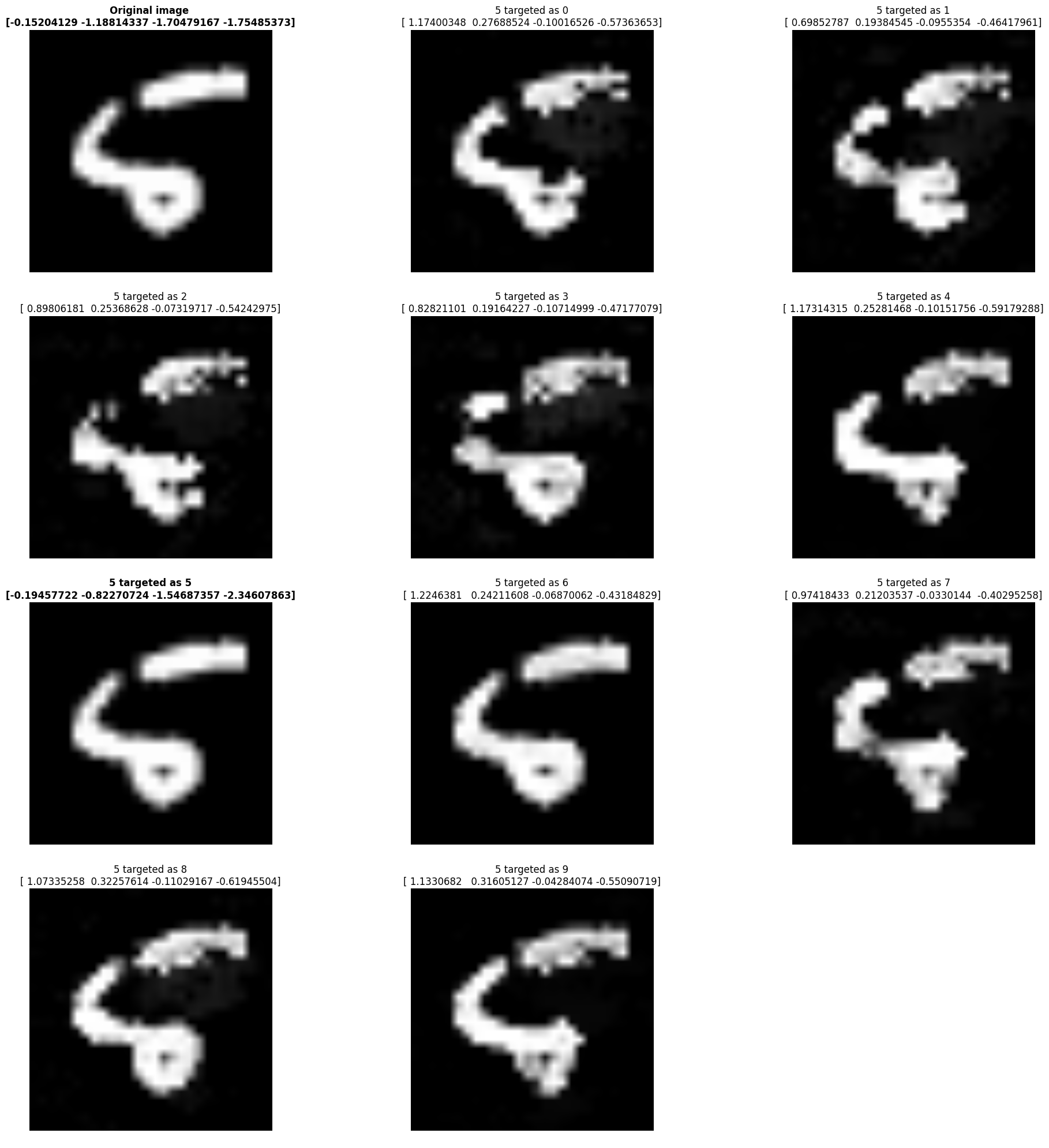}
\caption{Example of an image (Number 5) along with its ensemble of adversarial counterparts targeting different classes. The Lyapunov exponents are as shown in the titles of the constituent images}
\label{ex_5_mnist}
\end{figure*}

The lyapunov exponents of the flattened images showcased in fig \ref{cw_mnist} were numerically computed with the \texttt{lyap\_e()} method implemented in \cite{nolds} with the following parameterization:
\begin{verbatim}
PARAM_MNSIT={emb_dim=10, matrix_dim=4, 
min_nb=min(2 * matrix_dim, matrix_dim + 4), 
min_tsep=0, tau=1}
\end{verbatim}
An example images along with its flattened \textit{quasi-time-series} representation and the computed Lyapunov exponents is shown in fig \ref{qts_mnist}. Notice the change of sign of the lyapunov exponents between the true image and the adversarial images. In time-series analysis, the existence of at least one positive Lyapunov exponent is interpreted as a strong indicator for chaos. As we introduce more adversarial noise, the prevalence of positive Lyapunov exponents becomes more prevalent. Fig \ref{ex_5_mnist}, contains an example of an image (digit-5) and all its 9 associated adversarially perturbed counterparts (each targeting a different class) and their lyapunov exponents. 
\subsection{Unsupervised clustering of input images based on Lyapunov exponents}
\begin{figure}[ht]
\centering
\subfigure[An image (class-5- MNIST), its flattened quasi-time-series representation obtained by flattening]{
\includegraphics[width=3.25 in, height=2.8 in]{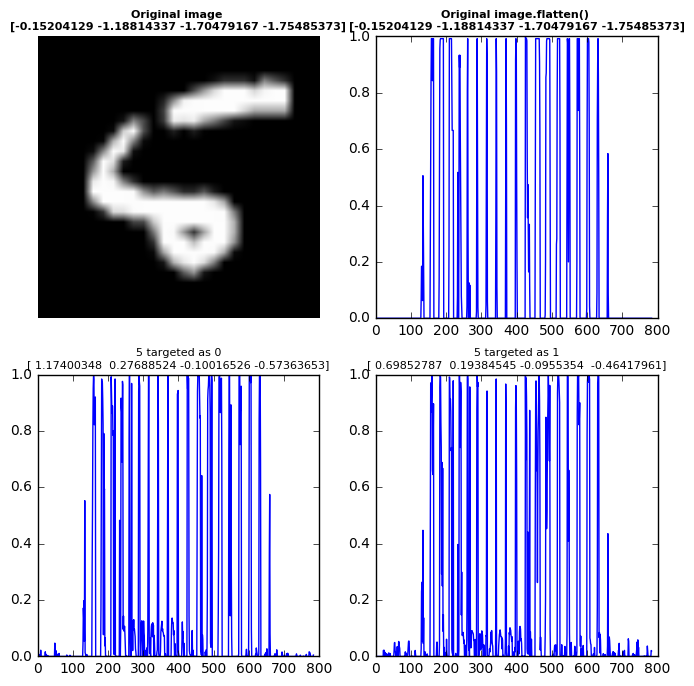}
    \label{qts_mnist}
}
\subfigure[2-D PCA embedding of the lyapunov exponents for the 100 images from the MNIST dataset shown in Fig \ref{cw_mnist}]{
\includegraphics[width=3.25 in, height=2.8 in]{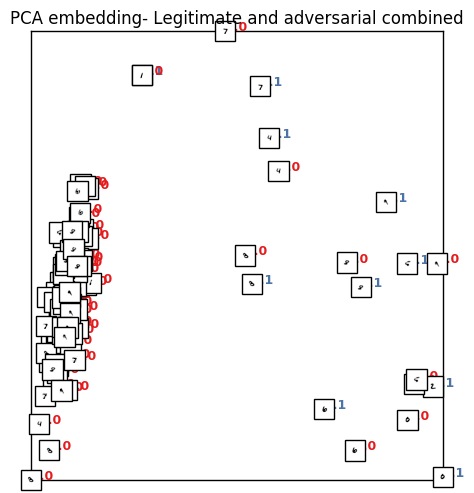}
\label{pca_mnist}
}
\caption[]{Quasi-time-series representations of an example image and their Lyapunov exponents}
\label{flatten_lyapunov}
\end{figure}

In fig \ref{pca_mnist}, where we have showcased the scatter-plot of the first 2 lyapunov exponents, we see clear clustering between the legitimate examples and the adversarial examples. We see that the only adversarial examples that are inside the cluster of legitimate examples are those where the target class is the same as the true class. (These are images whose every element is just a fixed $\varepsilon$ offset from the true image)
\subsection{Supervised classification}
\begin{figure*}[!t]
\centering
\includegraphics[width=7.75 in, height=3.5 in] {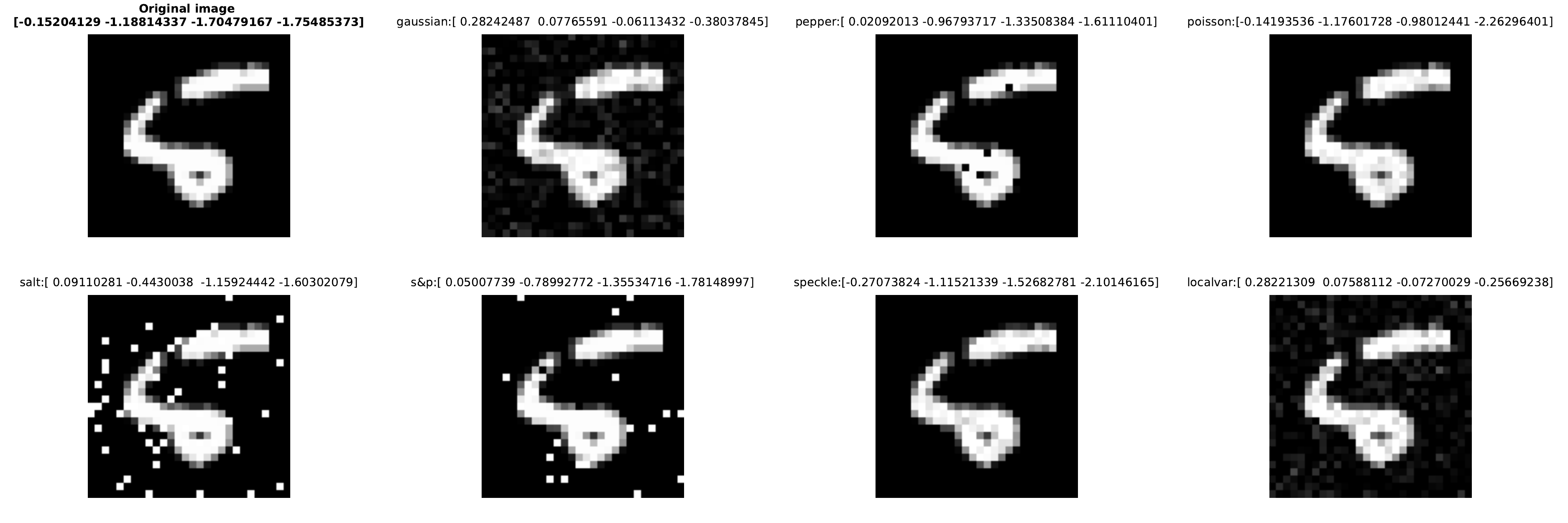}
\caption{The images of an example image sample (5) along with its noisy counterparts and their Lyapunov coefficients }
\label{qts_mnist_noise}
\end{figure*}

\begin{figure*}[ht]
\centering
\subfigure[Feature matrix heat-map]{
\includegraphics[width=3.25 in, height=2.8 in] {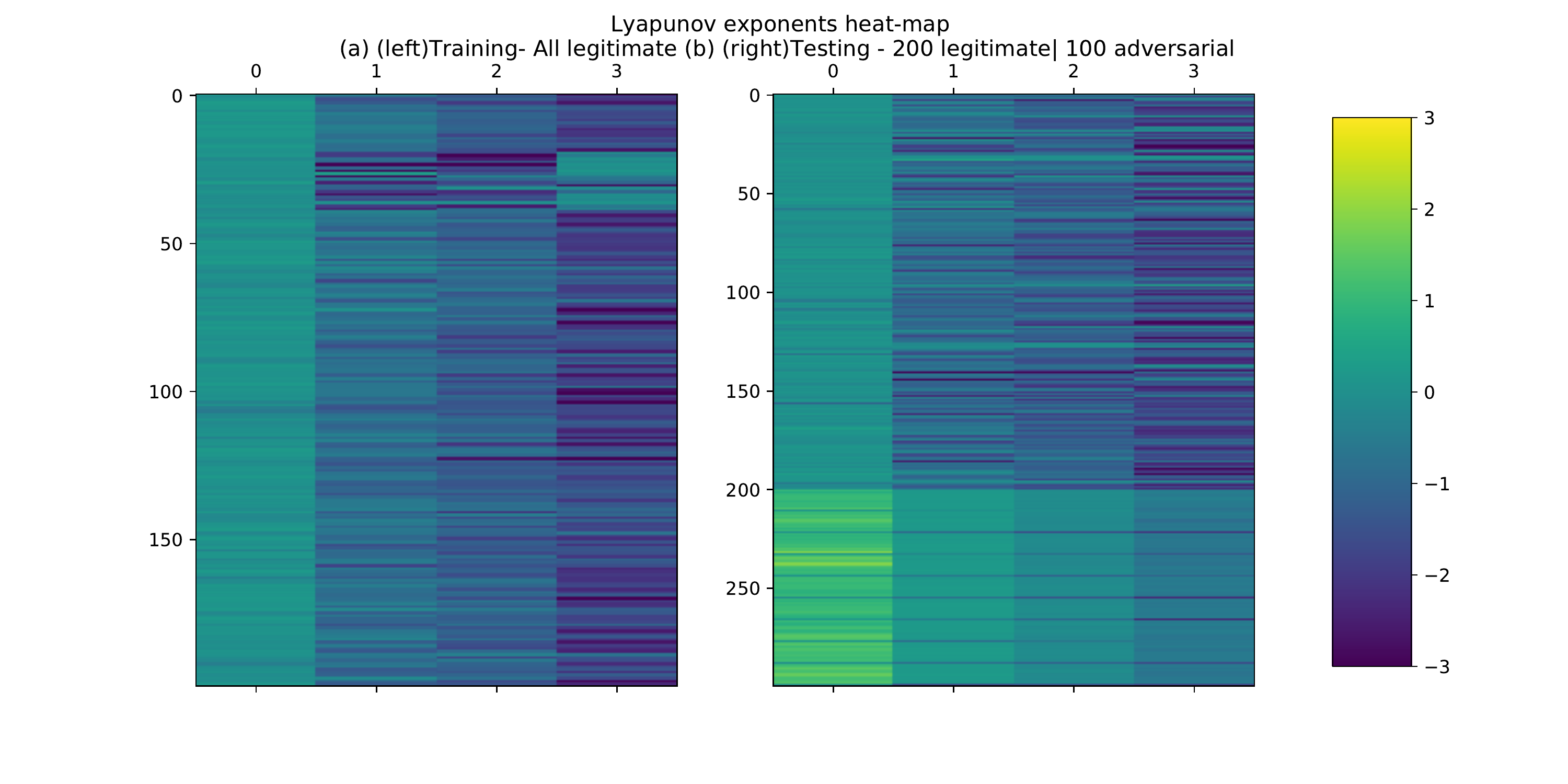}
    \label{f_lyap}
}
\subfigure[Iso-Forest scatter-plot]{
\includegraphics[width=3.25 in, height=2.8 in]{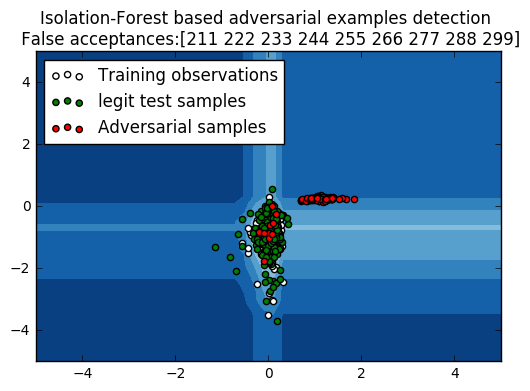}
    \label{iso_forest}
}

\end{figure*}
Detection can also be performing from the viewpoint of outlier detection where we \textit{train} a one-class classifier (ex: Isolation-Forest \cite{iso_forest}) with just the inlier (legitimate) images' lyapunov exponents used during training. Fig \ref{f_lyap} shows the heatmap of the features in the train and test subsets (200: Training samples| 300 Test samples of which 200 are legitimate and 100 are adversarial). We received $100\%$ attacker rejection rate and a $83.5\%$ true acceptance rate (or a $16.5\%$ False-alarm rate) with no feature engineering or hyper-parameter tuning.
Fig \ref{iso_forest} showcases the scatter-plot of the train and test samples along with the decision boundaries found by the isolation forest algorithm.
\subsection{Effect of non-adversarial noise on the detection performance}
In order to ascertain the effect of non-adversarial noise on the detection performance, we chose the following noise models \footnote{\url{http://scikit-image.org/docs/dev/api/skimage.util.html}}:
\begin{verbatim}
    ['gaussian','pepper','poisson','salt',
    'Salt and Pepper','Speckle','local-variance Gaussian']
\end{verbatim}
(Here 'local-variance Gaussian' refers to Gaussian-distributed additive noise with specified
local variance at each pixel of the image). 

\begin{figure}[!t]
\centering
\includegraphics[width=3.25 in, height=2.8 in]{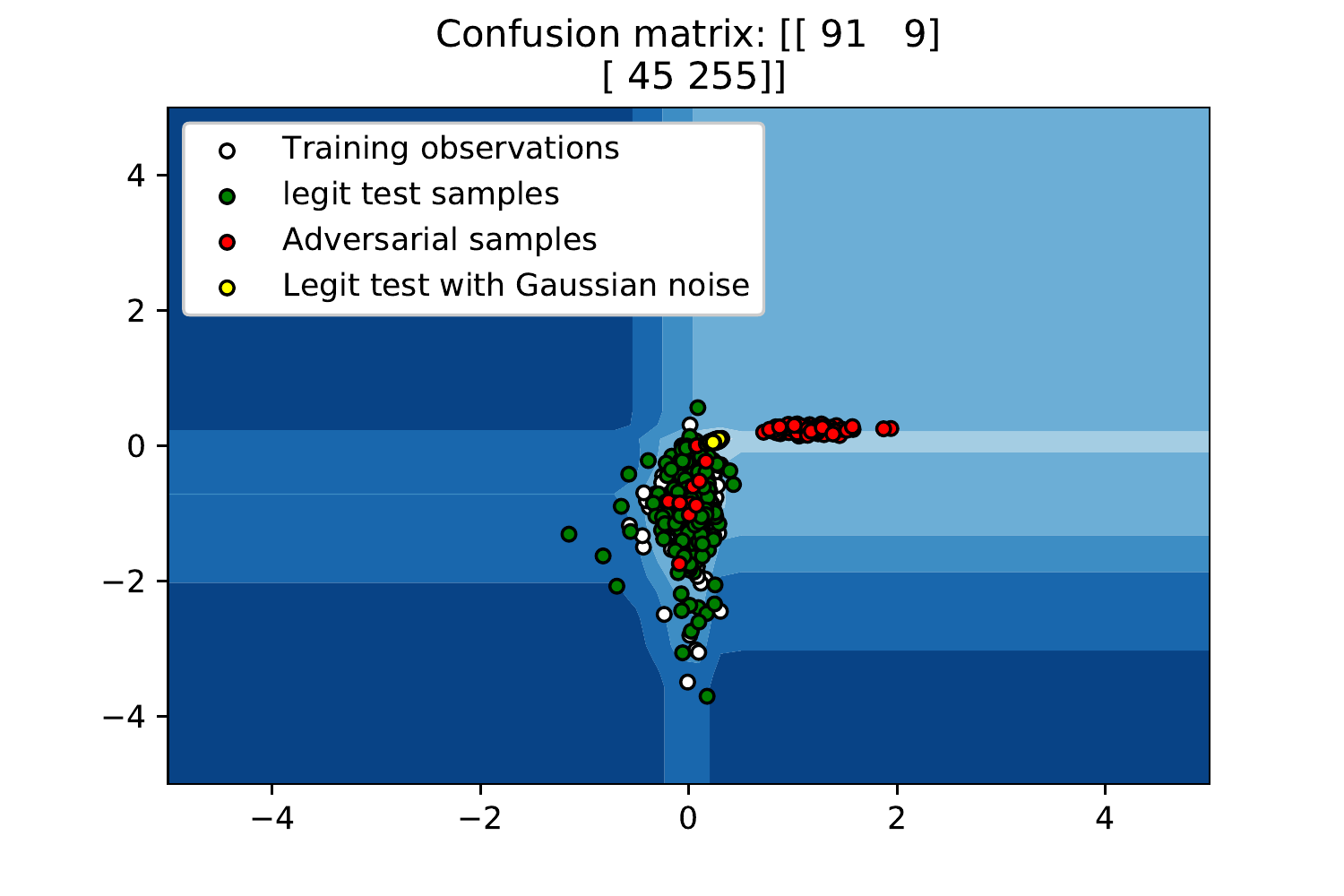}
\label{iso_forest_g}
\caption{Iso-Forest scatter-plot with Gaussian noise perturbed examples}
\end{figure}

The effect of these noise-models on the image and its lyapunov exponents are shown in fig \ref{qts_mnist_noise}. As seen, the 'local-variance Gaussian' noise has the most discernible effect on the lyapunov exponents and hence is chosen as the model of choice to perturb the legitimate images with. 
\subsection{Augmenting the dataset with non-adversarial perturbed images}
We added the training datasets with $100$ noisy images (with Gaussian perturbations) and re-train the \textit{isoforest} classifier. The results are as shown in fig \ref{iso_forest_g}. We see that the false alarm rate drops to $15\%$ while the attacker rejection rate is almost unchanged (only 1 attacker image is \textit{accepted in}).

We further observe that when the magnitude of random noise is selected from the distribution of norm distances among adversarial perturbations, the classifier retains its ability to distinguish between adversarially perturbed and randomly perturbed images.

\subsection{Testing the defense with other attacks}
We consider whether the Lyapunov method can be used to detect adversarial perturbations from attacks other than the Carlini-Wagner-$l2$ attack. We consider the Fast Gradient Sign Method \cite{FGSM}, the Jacobian Saliency Map Attack \cite{JSMA}, DeepFool \cite{DeepFool}, and the attack presented by Madry et al \cite{Madry}. We consider both the targeted and untargeted versions of each attack, where applicable\footnote{DeepFool does not have a targeted variant.}. We use the default parameters of the CleverHans library wherever possible. Where CleverHans does not provide a default value, we use the values referenced in the original paper describing the attack.

We find that when trained on only the first two Lyapunov exponents, the isolation forest does poorly on most of the attacks. However, accuracy improves significantly when we train on four-dimensional data. In both cases, we train using natural MNIST images as the inlier set. The true negative rates for these two classifiers on data generated by each of the attacks are presented in Table \ref{table:other-attacks}.

\begin{table}[ht]
	\centering
	\begin{tabular}{|c||c|c|c|}
		\hline
		Attack & Targeted/Untargeted & 2D & 4D \\ \hline\hline
		Carlini Wagner   & Targeted    & 0.90  & 0.91 \\ \hline
		& Untargeted  & 1.0   & 1.0  \\ \hline
		FGSM      		 & Targeted    & 0.0   & 0.62 \\ \hline
		& Untargeted  & 0.0   & 0.8  \\ \hline
		JSMA      		 & Targeted    & 0.03  & 0.04 \\ \hline
		& Untargeted  & 0.03  & 0.04 \\ \hline
		Madry et al.     & Targeted    & 0.0   & 1.0  \\ \hline
		& Untargeted  & 0.0   & 1.0  \\ \hline
		DeepFool         & Targeted    & NA    & NA   \\ \hline
		& Untargeted  & 1.0   & 1.0  \\ \hline
	\end{tabular}
	\label{table:other-attacks}
	\caption{True positive rates for several targeted and untargeted attacks.}
\end{table}

\subsection{Testing the effect of a new attack}
In previous sections, we used a 1-class Isolation Forest classifier to learn an inlier set of unmodified images. In this experiment, we test whether a classifier trained on both positive and negative data generated using known adversarial attacks can outperform the 1-class classifier.

\begin{figure*}[!t]
\centering
\includegraphics[width=3in] {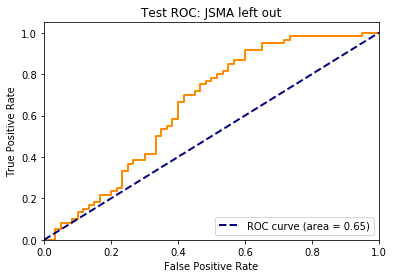}
\includegraphics[width=3in] {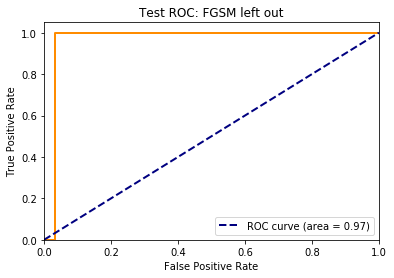}
\includegraphics[width=3in] {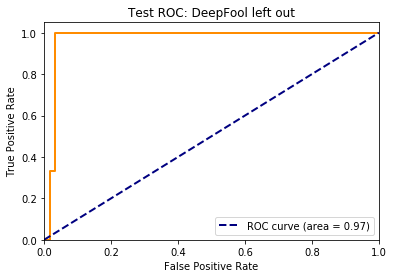}
\includegraphics[width=3in] {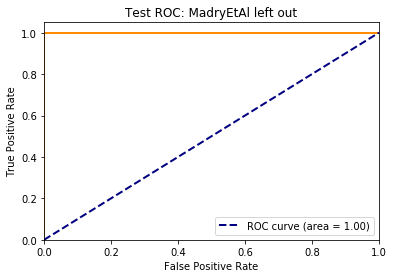}
\includegraphics[width=3in] {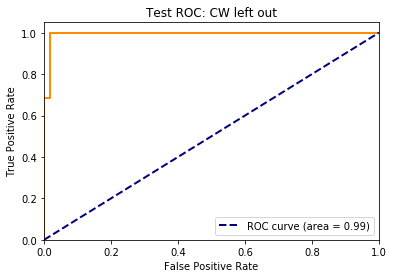}
\caption{ROC curves for logistic models trained on all but one attack and tested on the left out attack.}
\label{fig:left-out-roc}
\end{figure*}

We train a logistic regression model on unmodified MNIST images and data generated using all but one of the untargeted attacks from the previous section. We then evaluate the model on a validation set consisting of natural images and images modified using the left-out attack. We find that the logistic model is able to achieve near-perfect performance on four out of five attacks. The model only fails to perform on data generated using JSMA, achieving an AUROC score of 0.61. On all other attacks, the model reaches an AUROC score between 0.97 and 1.0. Exact AUROC scores and ROC curves are shown in Figure \ref{fig:left-out-roc}.

\section{Conclusion and future work}
Via this paper, we have sought to disseminate a serendipitous discovery entailing usage of lyapunov exponents as a model-agnostic tool that can be used to pre-filter input images as potentially adversarially perturbed. We have shown the validity of the idea \textit{defensing} against images that were adversarially perturbed using the Carlini-Wager-$l_2$ attack procedure across 2 datasets, namely MNIST and fashion-MNIST.
\\ We have used the latest version of CleverHans library (version : \texttt{2.0.0-7f7f9b18a1988fdf6d37d5c40deabae6}) and have open-sourced the code to ensure repeatability of the results presented here. 
\\We are currently investigating its potential across various other datasets and attacks.

\begin{figure*}[!t]
\centering
\includegraphics[width=7 in, height=8 in]{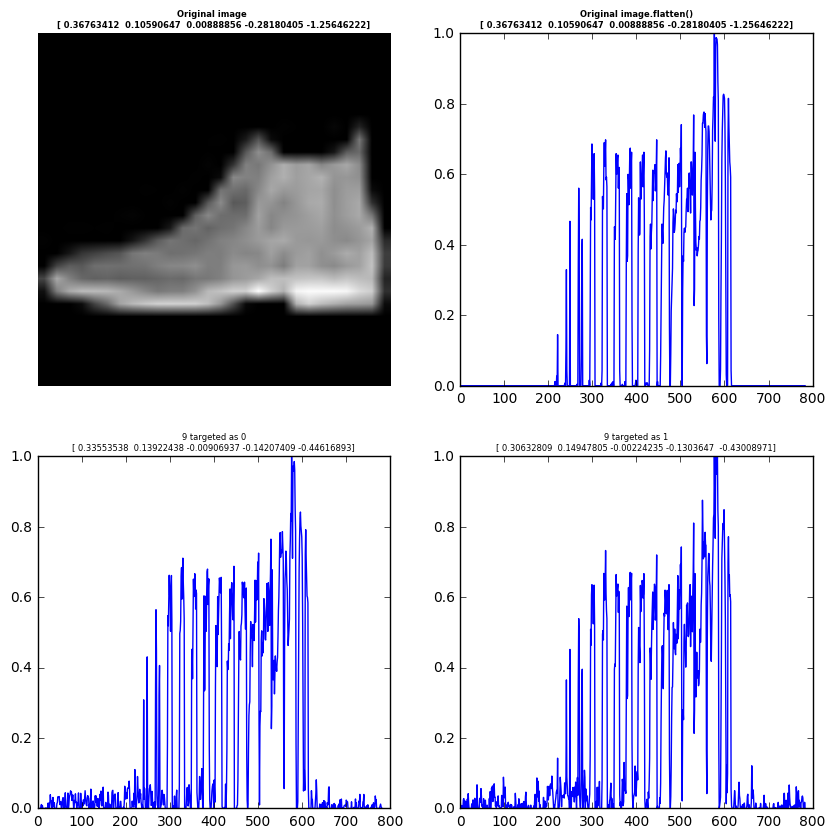}
\caption{2-D PCA embedding of the lyapunov exponents for the 100 images from the Fashion-MNIST dataset }
\label{flatten_lyapunov}
\end{figure*}

\begin{figure*}[!t]
\centering
\includegraphics[width=7 in, height=8 in]{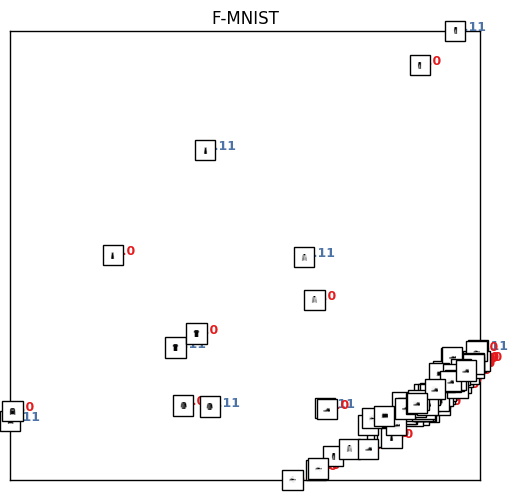}
\caption{2-D PCA embedding of the lyapunov exponents for the 100 images from the Fashion-MNIST dataset}
\label{pca_fmnist}
\end{figure*}

\begin{figure*}[!t]
\centering
\includegraphics[width=7 in, height=8 in]{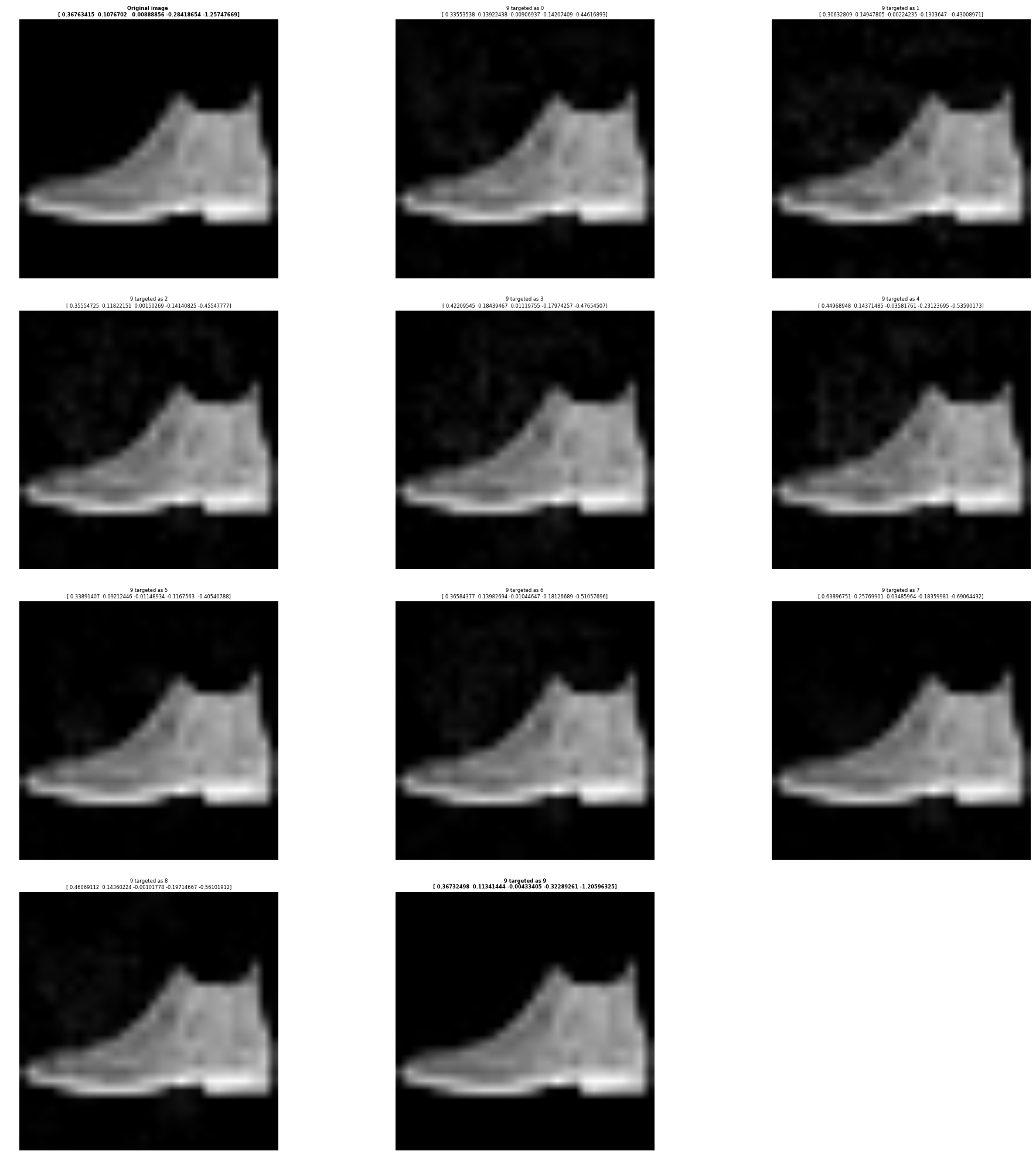}
\caption[Optional caption for list of figures]{Example of an image (Shoe) along with its ensemble of adversarial counterparts targeting different classes. The Lyapunov exponents are as shown in the titles of the constituent images}
\label{ex_5_fmnist}
\end{figure*}

\bibliographystyle{abbrv}
\bibliography{sigproc}  
%

\balancecolumns 
\end{document}